\documentclass[runningheads]{llncs}
\usepackage{epsfig}
\usepackage{subfigure}
\usepackage{wrapfig}
\usepackage{url}
\usepackage{psfrag}

\begin{document}

\pagestyle{headings}
\mainmatter

\title{Computational Complexity and Simulation of Rare Events of Ising Spin Glasses}
\titlerunning{Computational Complexity of Spin Glasses}

\author{Martin Pelikan\inst{1}, Jiri Ocenasek\inst{2}, Simon Trebst\inst{2}, Matthias Troyer\inst{2}, \\and Fabien Alet\inst{2}}
\authorrunning{Martin Pelikan et al.}

\institute{Dept. of Math. and Computer Science, 320 CCB\\
University of Missouri at St. Louis\\
8001 Natural Bridge Rd.,
St. Louis, MO 63121\\
\email{pelikan@illigal.ge.uiuc.edu}
\vspace*{1ex}
\and
Computational Laboratory (CoLab)\\
Swiss Federal Institute of Technology (ETH)\\
CH-8092 Z\"{u}rich, Switzerland\\
\email{jirio@inf.ethz.ch}\\
\email{\{trebst,troyer,falet\}@comp-phys.org}}

\maketitle


\begin{abstract}
\begin{sloppy}
We discuss the computational complexity of random 2D Ising spin
glasses, which represent an interesting class of constraint
satisfaction problems for black box optimization. Two extremal
cases are considered: (1) the $\pm J$ spin glass, and (2) the
Gaussian spin glass. We also study a smooth transition between
these two extremal cases. The computational complexity of all
studied spin glass systems is found to be dominated by rare events
of extremely hard spin glass samples. We show that complexity of
all studied spin glass systems is closely related to Fr\'echet
extremal value distribution. In a hybrid algorithm that combines
the hierarchical Bayesian optimization algorithm (hBOA) with a
deterministic bit-flip hill climber, the number of steps performed
by both the global searcher (hBOA) and the local searcher follow
Fr\'echet distributions. Nonetheless, unlike in methods
based purely on local search, the parameters of these
distributions confirm good scalability of hBOA with local search.
We further argue that standard performance measures for
optimization algorithms---such as the average number of
evaluations until convergence---can be misleading. Finally, our
results indicate that for highly multimodal constraint
satisfaction problems, such as Ising spin glasses,
recombination-based search can provide qualitatively better
results than mutation-based search.
\end{sloppy}
\end{abstract}


\section{Introduction}

The spin glass problem is an old-standing but still intensively
studied problem in physics~\cite{Mezard:86}. First, experimental
realizations of spin glass systems do exist and their properties,
in particular their dynamics, are still not well explained.
Second, spin glasses pose a challenging, unsolved problem in theoretical
physics since the nature of the spin glass state at low
temperatures is not understood. It is widely believed that this is
due to the intrinsic complexity of the rough energy landscape of
spin glasses.

In statistical physics, one usual goal is to calculate a desired
quantity (e.g. magnetization) over a distribution of
configurations of a spin glass system for a given temperature. The
probability of observing a specific spin configuration, $C$, of
the spin glass is governed by the Boltzmann distribution, that is
to say it is inversely proportional to the exponential of the
ratio of its energy and temperature : $p(C) \sim \exp(-E(C)/T)$.
Thus, as temperature decreases, the distribution of possible
configurations of the spin glass concentrates near the
configurations with minimum energy, which are also called ground
states. The ground-state properties capture most of the low
temperatures physics, and it is therefore very interesting to find
and study them.

From another perspective, spin glasses represent an interesting
class of problems for black-box optimization where the task is to
find ground states of a given spin glass sample, because the
energy landscape in most spin glasses exhibits features that make
it a challenging optimization benchmark. One of these features is
the large number of local optima, which often grows exponentially
with the number of decision variables (spins) in the problem.
Because of the large number of local optima, using local search
operators, such as mutation, is almost always intractable.

In this paper we present, analyze, and discuss a series of
experiments on 2D Ising spin glasses. Random spin glass instances
for a fixed lattice geometry (square lattice) are generated by
randomly sampling a fixed distribution of coupling constants. We
distinguish two basic classes of random 2D Ising spin glass
systems: (1) coupling constants are initialized
randomly to either $+1$ or $-1$, and (2) coupling constants are generated from a zero-mean Gaussian
distribution. A transition between these two cases is also
considered. We apply the hierarchical Bayesian optimization
algorithm (hBOA) with local search to all considered classes of
spin glasses, and provide a thorough statistical analysis of hBOA
performance on a large number of problem instances in each class.
The results are discussed in the context of state-of-the-art Monte
Carlo methods, such as the Wang-Landau algorithm \cite{Wang:01}
and the multicanonical method \cite{Berg:92}. Finally, we identify
important lessons from this work for genetic and evolutionary
computation.

In the following we present a short review of the hierarchical
Bayesian optimization algorithm and extremal value distributions
used in the statistical analysis. In
section~\ref{section-methodology} we define the 2D Ising spin
glass systems analyzed in this work, and introduce several
classes of random spin glass instances.
Section~\ref{section-experiments} presents experimental
methodology and results. Finally,
Section~\ref{section-conclusions} summarizes and concludes the
paper.


\section{Numerical methods and statistical analysis}
\label{section-background}

This section briefly discusses the hierarchical Bayesian
optimization algorithm (hBOA)~\cite{Pelikan:01*,Pelikan:03b} and
extremal value distributions, which will be used to analyze
experimental results.

\subsection{Hierarchical Bayesian optimization algorithm (hBOA)}
The hierarchical Bayesian optimization algorithm
(hBOA)~\cite{Pelikan:01*,Pelikan:03b} is one of the most advanced
genetic and evolutionary algorithms based primarily on selection
and recombination. hBOA evolves a population of candidate
solutions to a given problem. Using a {\em population} of
solutions as opposed to a single solution has several advantages;
for example, it enables simultaneous exploration of multiple
regions in the search space, it can help to alleviate the effects
of noise in evaluation, and it allows the use of statistical and
learning techniques to identify regularities in the black-box
optimization problem under consideration.

The first population of candidate solutions is usually generated
according to uniform distribution over all candidate solutions. The population is updated for a number of iterations
using two basic operators: (1) selection, and (2) variation. The
selection operator selects better solutions at the expense of the
worse ones from the current population, yielding a population of
promising candidates. The variation operator starts by learning a
probabilistic model of the selected solutions that encodes
features of these promising solutions and the inherent
regularities. hBOA uses Bayesian networks with local
structures~\cite{Chickering:97} to model promising solutions. The
variation operator then proceeds by sampling the probabilistic
model to generate new solutions. The new solutions are
incorporated into the original population using the restricted
tournament replacement (RTR)~\cite{Harik:95a}, which ensures that
useful diversity in the population is maintained over long periods
of time. A more detailed description of hBOA can be found
in~\cite{Pelikan:thesis}.

To improve candidate solutions locally, hBOA applies a
deterministic bit-flip hill-climber to each newly generated
candidate solution that improves the solution by single-bit flips
until no further improvement is possible. Flips that produce
better solutions are of higher priority. It was previously
shown that local search can significantly reduce population sizes
for various optimization problems, including the spin glass
problem~\cite{Pelikan:03*}.

\subsection{Extremal value distributions}
\label{section-evd}

Several quantities related to the computational complexity studied
in this work are found to follow extremal value distributions. The
central limit theorem for extremal values states that the extremes
of large samples are distributed according to one of three
extremal value distributions, depending on whether their shapes
are fat-tailed (tails decay polynomially), exponential (tails
decay exponentially), or thin-tailed (tails decay faster than
exponentially)~\cite{Fisher:28}. The integrated probability
density function for any of these extremal value distributions can
be written as
\begin{equation}
H_{\xi;\mu;\beta}(x) = \exp\left(-{\left( 1 + \xi \frac{x-\mu}{\beta}\right)}^{\frac{1}{\xi}}\right),
\end{equation}
where $\mu$ is the location parameter, $\beta$ is the scaling
parameter, and $\xi$ is the shape parameter that indicates how
fast the tail decays. If $\xi<0$, $H_{\xi;\mu;\beta}(x)$ represents the Fr\'echet
distribution (polynomial decay), if $\xi=0$ it represents the
Gumbel distribution (exponential decay), and if $\xi>0$ it
represents the Weibull distribution (faster than exponential
decay). Distributions encountered in this work are Fr\'echet
distributions, where the shape parameter $\xi$ determines the
power law decay of the fat tails of the distribution

\begin{equation}
\frac{dH_{\xi;\mu;\beta}}{dx}
\stackrel{x\to\infty}{-\!\!\!\!-\!\!\!\!-\!\!\!\!-\!\!\!\!\longrightarrow}
x^{-(1-1/\xi)} \;.
\label{eq:Tail}
\end{equation}

From this asymptotic behavior one can see that the $m$-th moment of a fat tailed
Fr\'echet distribution (with $\xi<0$) is well defined only if $|\xi| < 1/m$.


\section{The Ising spin glass}
\label{section-methodology}

A 2D spin glass system consists of a regular 2D grid containing
$N$ nodes which correspond to the spins. The edges in the grid
connect nearest neighbors. Additionally, edges between the first
and the last element in each dimension are added to introduce
periodic boundary conditions. 
for an example 2D spin glass structure consisting of $9$ spins
distributed on a $3\times 3$ square lattice.


With each edge there is a real-valued constant associated which
gives the strength of spin-spin coupling. For the classical Ising
model each spin can be in one of two states: $+1$ or $-1$. Each
possible set of values for all spins is called a spin
configuration. Given a set of (random) coupling constants,
$J_{i,j}$, and a configuration of spins, $C$, the energy can be
computed as
\begin{equation}
E(C) = \sum_{\langle i,j\rangle} s_i J_{i,j} s_j \;,
\end{equation}
where $i,j \in\{0, 1, \ldots, N-1\}$ denote the spins (nodes) and
$\langle i,j\rangle$ nearest neighbors on the underlying grid
(allowed edges). The random spin-spin coupling constants $J_{i,j}$ for a
particular spin glass instance are given on input.

In statistical physics, the usual task is to integrate a known
function over all possible configurations of spins,
 where the configurations are distributed
according to the Boltzmann distribution. Probability of
encountering a configuration, $C$ at temperature $T$ is given by
\begin{equation}
\label{eq-boltzmann-distribution}
p(C) = \frac{\exp\left({-E(C)/T}\right)}{\sum_{\tilde{C}} \exp\left({-E(\tilde{C})/T}\right)} \;.
\end{equation}

From the physics point of view, it is interesting to know the ground states (configurations
associated with the minimum possible energy). Finding extremal energies
then corresponds to sampling the Boltzmann distribution with temperature
approaching $0$ and thus the problem of finding ground states is simpler {\it
  a priori} than integration over a wide range of temperatures. However, most
of the conventional methods based on sampling the above Boltzmann distribution \ref{eq-boltzmann-distribution} fail to find the ground states configurations because they get often trapped in a local minimum.

The problem of finding ground states is a typical optimization problem, where the task is to find an optimal
configuration of spins that minimizes energy. Although
polynomial-time deterministic methods exist for both types of 2D
spin glasses~\cite{Galluccio:99,Galluccio:99a}, most algorithms
based on local search operators, including a (1+1) evolution
strategy, conventional Monte Carlo simulations, and Monte Carlo
simulations with Wang-Landau~\cite{Wang:01} or multicanonical
sampling~\cite{Berg:92},
scale exponentially and are thus impractical for solving this
class of problems. The origin for this slowdown is due to the
suppressed relaxation times in the Monte Carlo simulations in the
vicinity of the extremal energies because of the enormous number
of local optima in the energy landscape. Recombination-based
genetic algorithms succeed if recombination is performed in a way
that interacting spins are located close to each other in the
representation; $k$-point crossover with a rather small $k$ can
then be used so that the linkage between contiguous blocks of bits
is preserved (unlike with uniform crossover, for instance).
However, the behavior of such specialized representations and
variation operators cannot be generalized to similar slowly
equilibrating problems which exhibit different energy landscapes,
such as protein folding or polymer dynamics.

In order to obtain a quantitative understanding of the disorder in
a spin glass system introduced by the random spin-spin couplings,
one generally analyzes a large set of random spin glass instances
for a given distribution of the spin-spin couplings. For each spin
glass instance the optimization algorithm is applied and results
statistically analyzed to obtain a measure of computational
complexity. Here we first consider two types of initial spin-spin
coupling distributions, the $\pm J$ spin glass and the Gaussian
spin glass.

\subsection{The $\pm J$ spin glass}
For the $\pm J$ Ising spin glass, each spin-spin coupling constant
is set randomly to either $+1$ or $-1$ with equal probability (see
lower right panel in Figure \ref{fig-transition-distribution}).
Energy minimization in this case can be transformed into a
constraint satisfaction problem, where the constraints relate
spins connected by a coupling constant. If $J_{i,j}>0$, then the
constraint requires spins $i$ and $j$ to be different, whereas if
$J_{i,j}<0$, then the constraint requires spins $i$ and $j$ to be
the same. Energy is minimized when the number of satisfied
constraints is maximized.

\subsection{Gaussian spin glasses}
In the Gaussian spin glass, coupling constants are generated
according to a zero-mean Gaussian distribution with variance one
(see upper left panel in Figure
\ref{fig-transition-distribution}). For real-valued couplings,
energy minimization can be casted as a constraint satisfaction
problem with weighted constraints.

\subsection{Transition between $\pm J$ and Gaussian spin glasses}
\begin{figure}[t]
\begin{center}
\epsfig{file=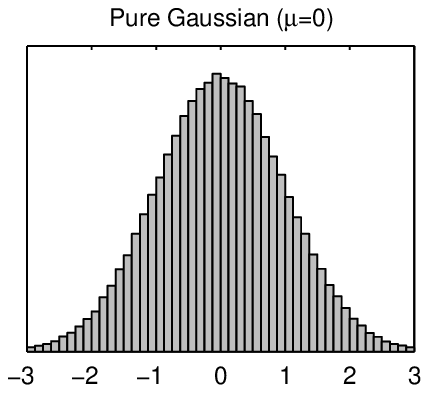,width=1.1in}~~~
\epsfig{file=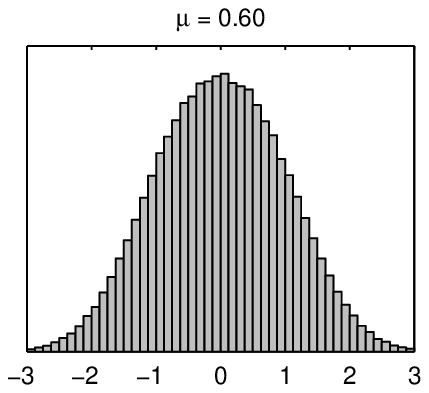,width=1.1in}~~~
\epsfig{file=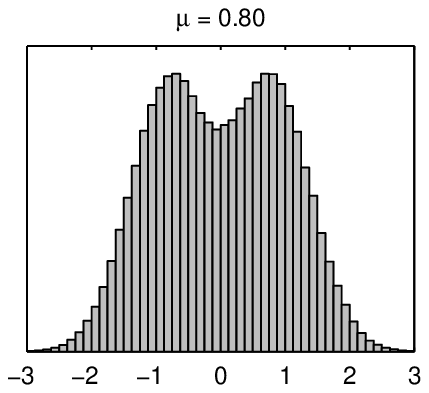,width=1.1in}\\~\\
\epsfig{file=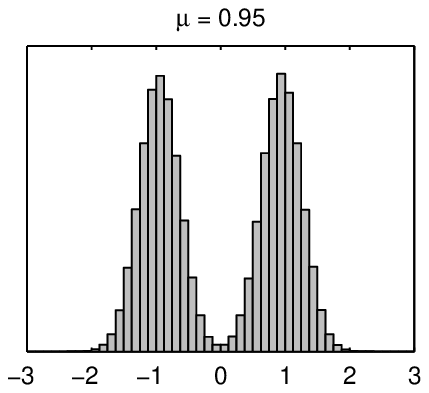,width=1.1in}~~~
\epsfig{file=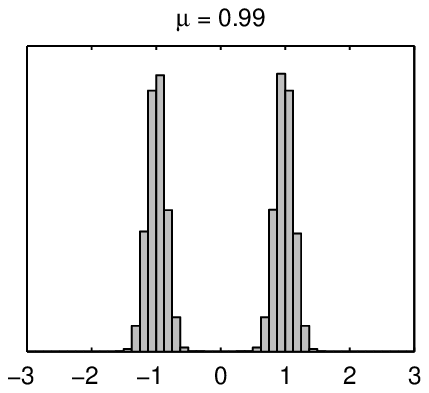,width=1.1in}~~~
\epsfig{file=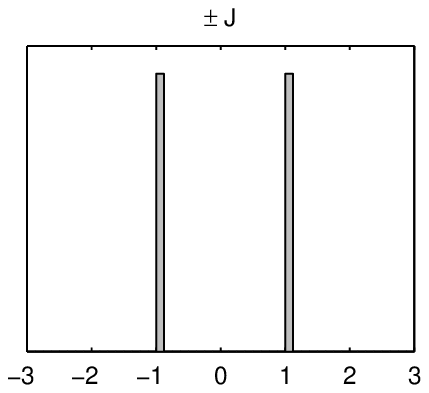,width=1.1in}
\end{center}
\vspace*{-3ex}
\caption{Distribution of coupling constants for the transition from the Gaussian (upper left) to the $\pm J$ spin glass (lower right).}
\label{fig-transition-distribution}
\end{figure}

To describe a smooth transition between the $\pm J$ and the
Gaussian spin glass we vary the distribution of spin-spin coupling
constants by defining a distribution as the sum of two Gaussian
distributions, described by means, $\pm \tilde{\mu}$, and
variance, $\tilde \sigma$, in such a way that the overall mean
becomes $\mu = 0$ and the  overall variance $\sigma = 1$. The
explicit form of the two Gaussians is thus given by
$\tilde{\sigma}^2=1-\tilde{\mu}^2$. The $\pm J$ spin glass
($\tilde{\mu} = 1$) and the Gaussian spin glass ($\tilde{\mu} =
0$) then describe the extremal cases of this new family of
distributions. The transition between the two extrema is then
described by varying $\tilde{\mu}$ between 0 and 1 which is
illustrated in Figure~\ref{fig-transition-distribution} for
$\tilde{\mu} = 0, 0.60, 0.80, 0.95, 0.99, 1$.

\section{Numerical experiments}
\label{section-experiments}

In the following we describe the numerical experiments in more detail and present results for the spin glasses described above.

\subsection{Description of experiments}

For $\pm J$ and Gaussian 2D spin glasses, systems with equal
number of spins in each dimension were used of size from
$n=8\times 8$ to $n=20\times 20$. For each system size, 1000
random samples were generated. hBOA with the deterministic local
searcher was then applied to find the ground state for each
sample. For the transition from $\pm J$ to Gaussian spin glasses,
we focused on a single system size, $n=10\times 10$.

For each spin glass sample, the population size in hBOA is set to
the minimum population size required to find the optimum in 10
independent runs. The minimum population size is determined using
bisection. The width of the final interval in bisection is at most $10\%$ of its
higher limit. Binary tournament selection without replacement is
used. The windows size in RTR is set to the number of spins of the
system under consideration, but it is always at most equal to
$5\%$ of the population size. The $5\%$ cap on the window size is
important to ensure fast convergence with even small populations.
The cap explains the difference between the results presented here
and the previous results, because populations are usually very
small for hBOA with local search on Ising spin
glasses~\cite{Pelikan:thesis}.

Performance of hBOA was measured by (1) $E_G$, the total number of
spin glass system configurations examined by hBOA (the number of
restarts of the local searcher), and (2) $E_L$, the total number
of steps of the local hill climber. Due to the lack of space, we
only analyze $E_G$. $E_L$ was greater than $E_G$ by a factor of
approximately $O(\sqrt{n})$. Clearly, we can expect that
$E_G<E_L$. Nonetheless, it is computationally much less expensive
to perform a local step in the hill climber than to evaluate a new
spin glass configuration sampled by hBOA.

\subsection{Results for $\pm J$ and Gaussian couplings}

\begin{figure}[t]
\begin{center}
\epsfig{file=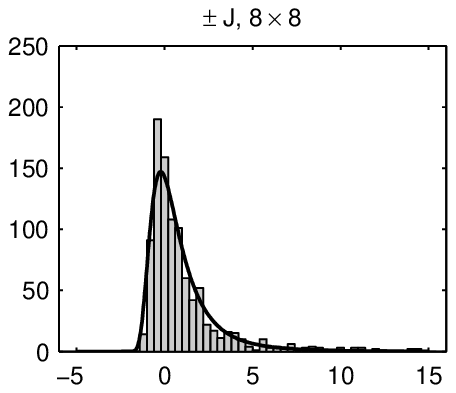,height=1.10in}~~~
\epsfig{file=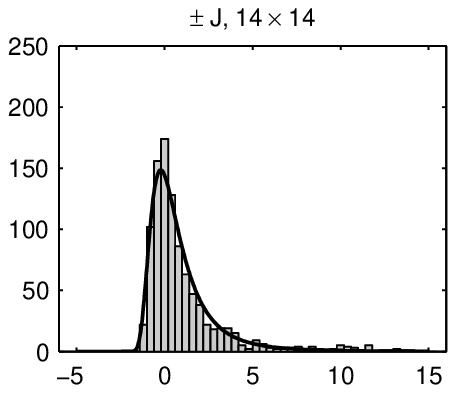,height=1.10in}~~~
\epsfig{file=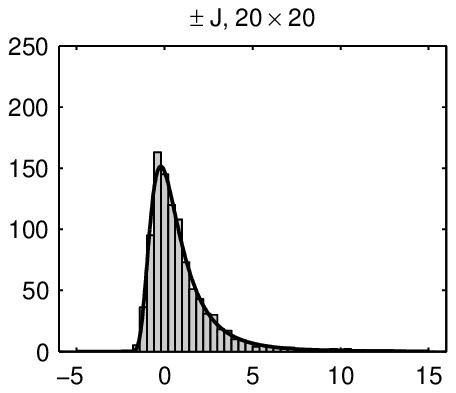,height=1.10in}\\~\\
\epsfig{file=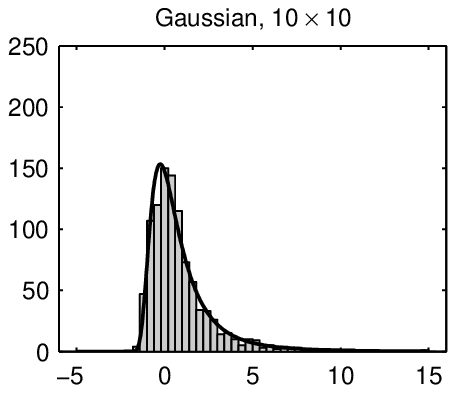,height=1.10in}~~~
\epsfig{file=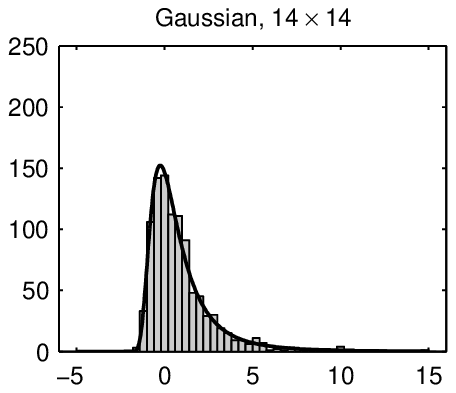,height=1.10in}~~~
\epsfig{file=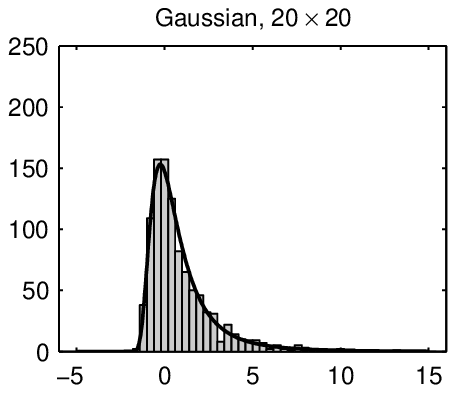,height=1.10in}
\end{center}
\vspace*{-3ex}
\caption{Distribution of $E_G$ for $\pm J$ spin glass systems of
varying size. $E_G$ and the density function are normalized using
$\mu$ and $\beta$.}
\label{figure-Frechet}
\end{figure}

The first important observation is that the distributions of $E_G$
and $E_L$ for all problem sizes and distributions of coupling
constants follow Fr\'echet extremal value distributions. Applying
a maximum likelihood estimator we can determine the parameters
$\mu$, $\beta$, and $\xi$ of these distributions defined in
Equation (1). Figure~\ref{figure-Frechet} shows the histograms and
the corresponding probability density function for $E_G$ for $\pm
J$ spin glasses of various sizes.

\begin{figure}[t]
\begin{center}
\epsfig{file=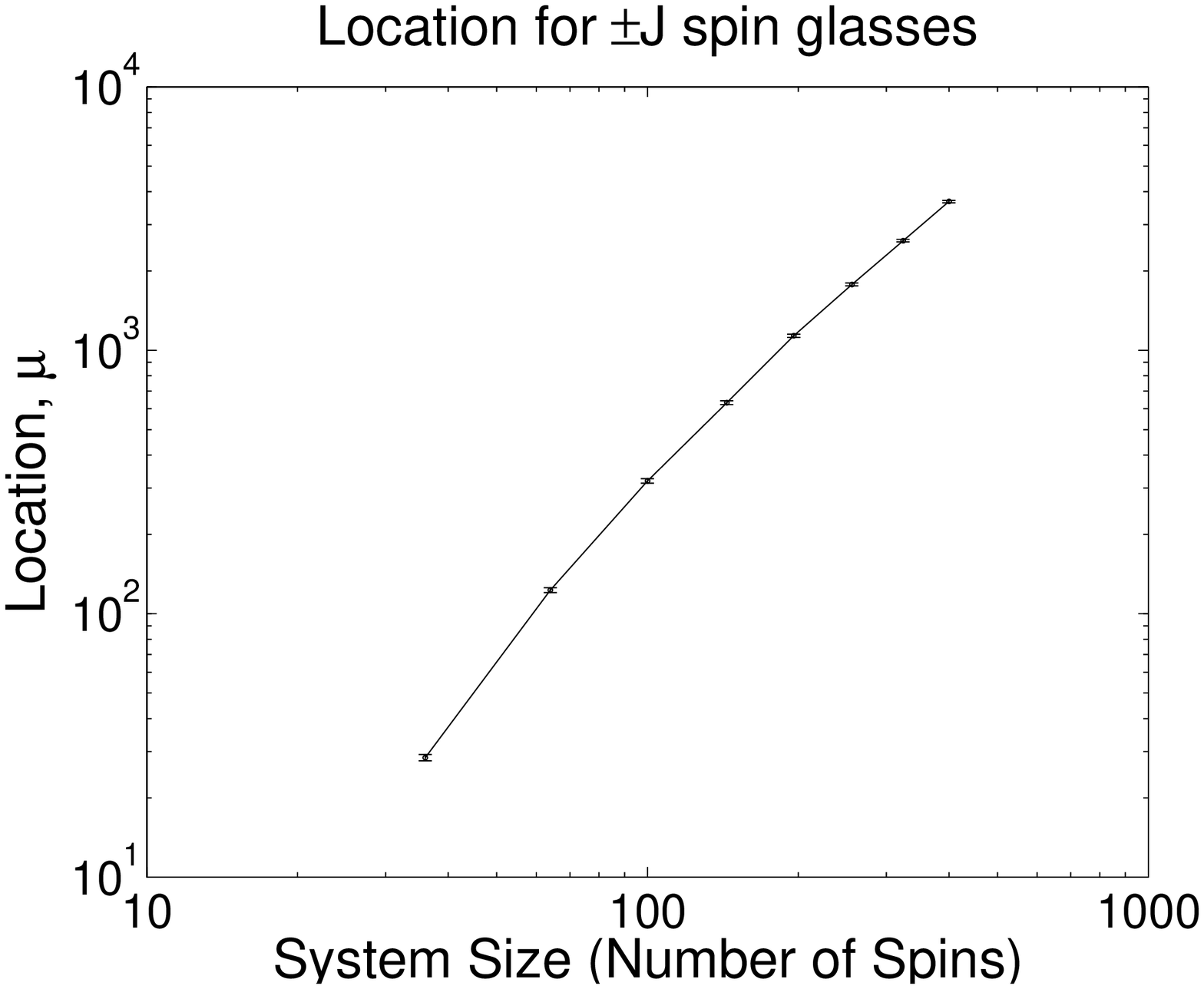,width=0.35\textwidth}~~~
\epsfig{file=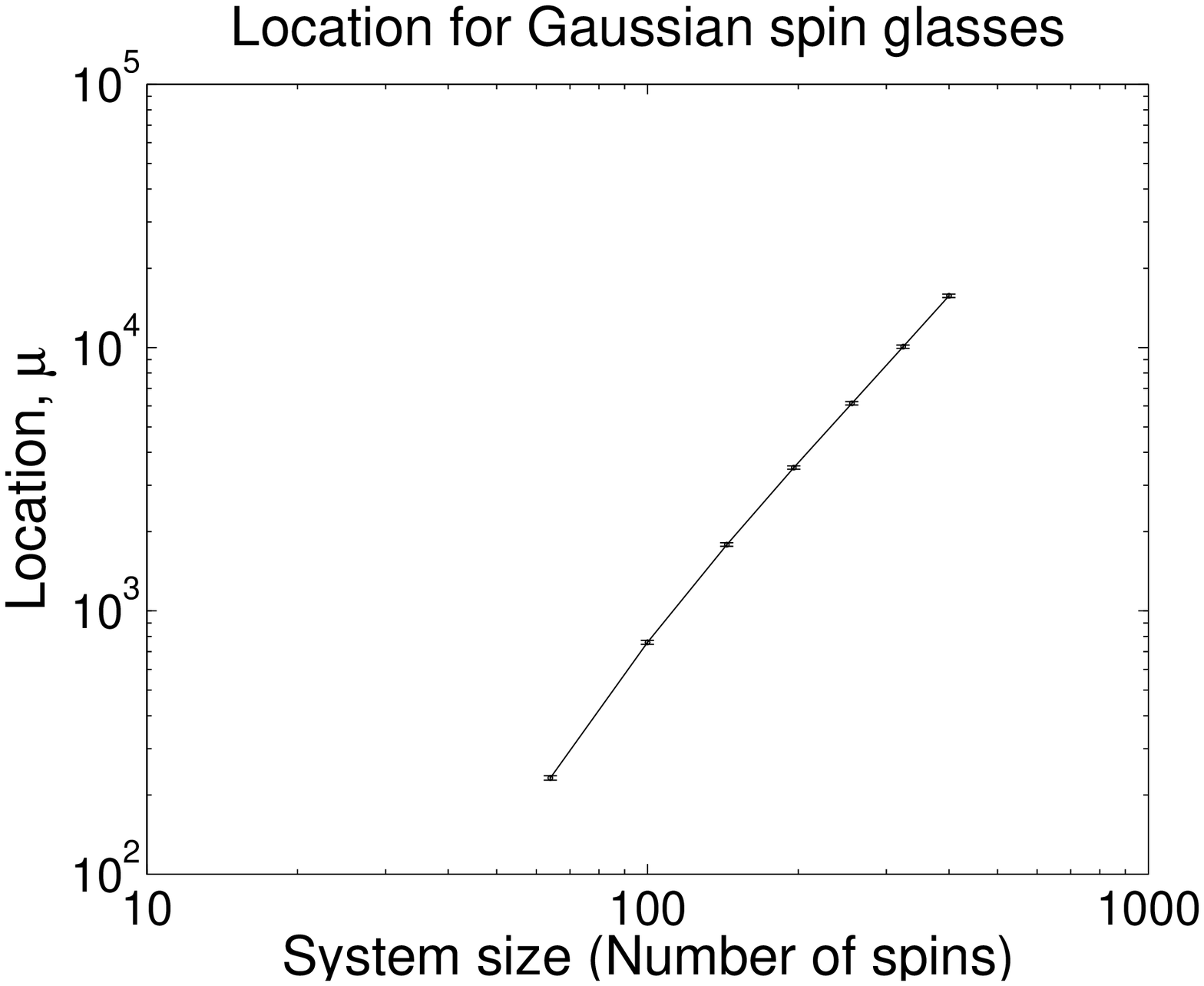,width=0.35\textwidth}\\~\\
\epsfig{file=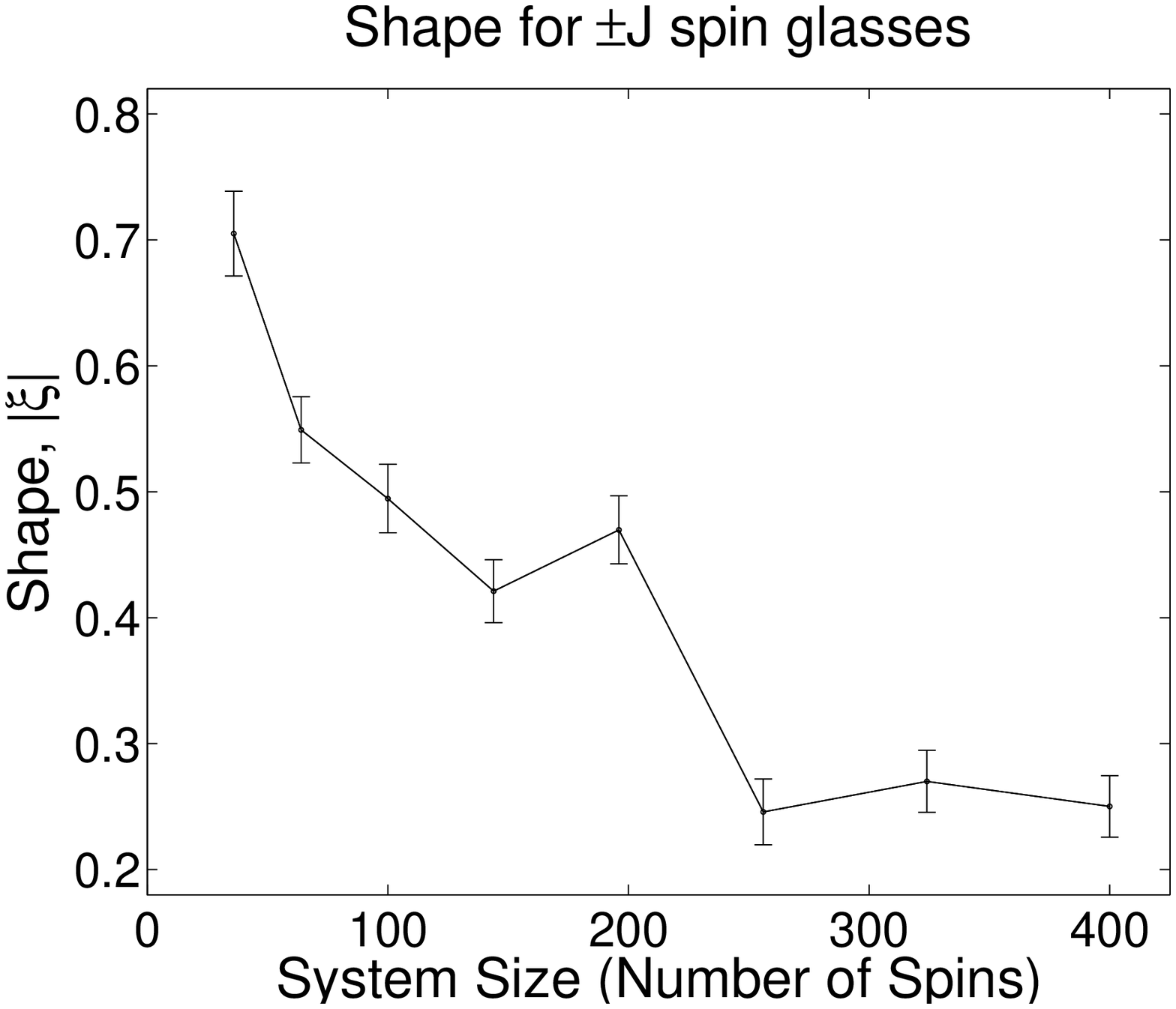,width=0.35\textwidth}~~~
\epsfig{file=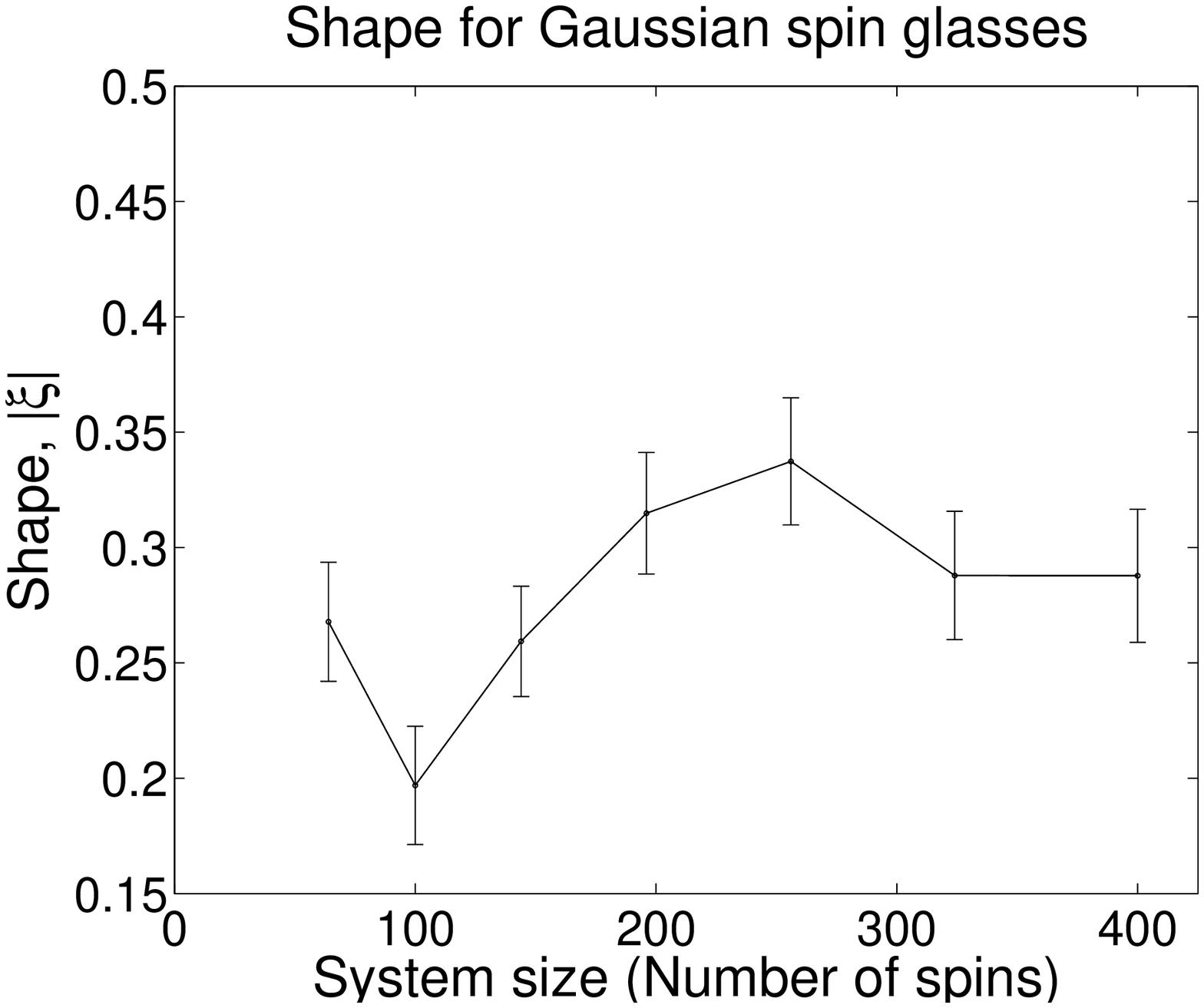,width=0.35\textwidth}
\end{center}
\vspace*{-3ex}
\caption{Location $\mu$ and shape $\xi$ for $\pm J$ and Gaussian
spin glasses using maximum likelihood estimation. Standard error
of the estimations displayed with error bars.}
\label{fig-location-and-shape}
\end{figure}

The location parameter $\mu$ indicating the most likely value of
$E_G$ can be used to determine the scalability of hBOA. In
Figures~\ref{fig-location-and-shape}a and
\ref{fig-location-and-shape}b the location parameter for both $\pm
J$ and Gaussian spin glasses are shown versus the system size.
Double logarithmic plots confirm that the location has an upper
polynomial bound. For the $\pm J$ spin glass, the order of that
polynomial approaches $1.5$ as system size $n$ grows, whereas for
Gaussian couplings, the order of the polynomial seems to approach
$2.2$.

Figure~\ref{fig-location-and-shape}c and
~\ref{fig-location-and-shape}d show the shape $\xi$ for both $\pm
J$ and Gaussian spin glasses with respect to the system size.
Since it is always smaller than 1, we conclude that the mean is
well-defined for all cases. For the variance (2nd moment) we find
the shape parameter to be smaller than 1/2 only for systems larger
than $n=10\times10$. Thus, for system smaller than $n=10\times10$
the variance is not well-defined and the mean has an infinite
error.

\subsection{Results for the transition between $\pm J$ and Gaussian couplings}

\begin{figure}[t]
\begin{center}
\epsfig{file=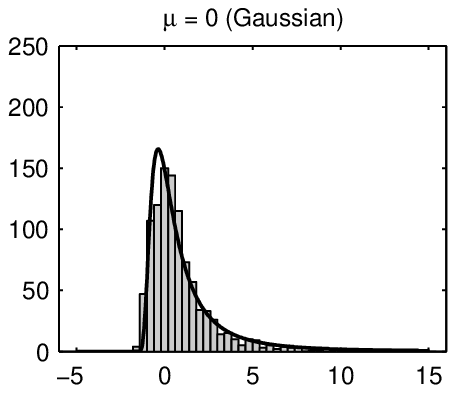,height = 1.1in}~~~
\epsfig{file=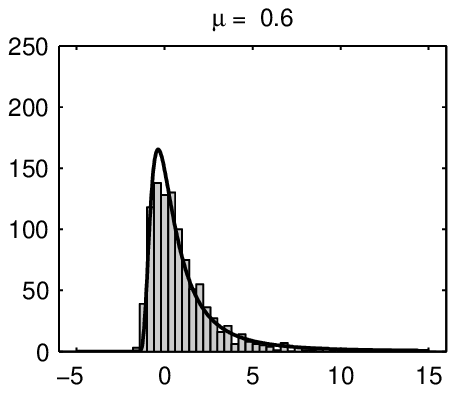,height = 1.1in}~~~
\epsfig{file=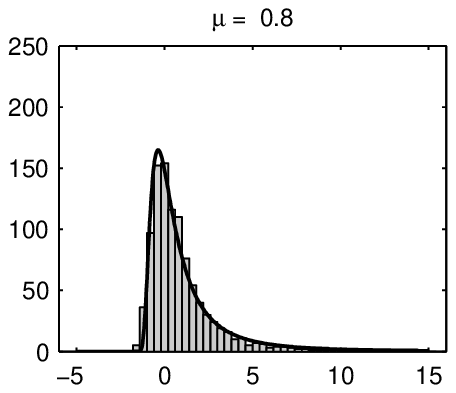,height = 1.1in}\\~\\
\epsfig{file=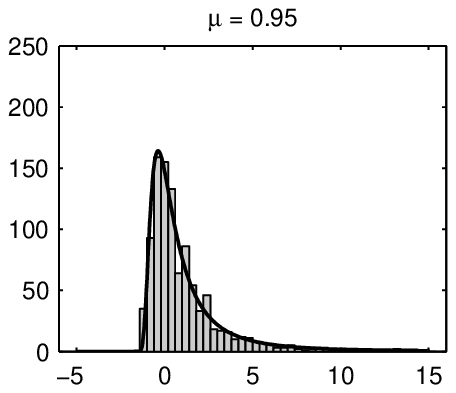,height = 1.1in}~~~
\epsfig{file=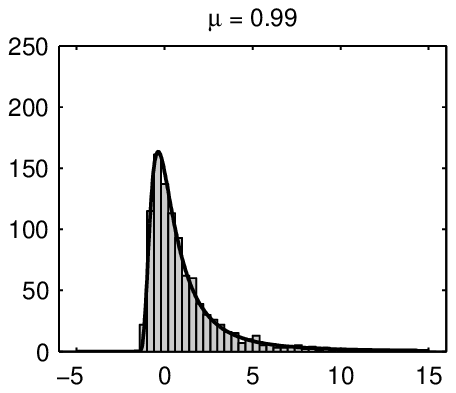,height = 1.1in}~~~
\epsfig{file=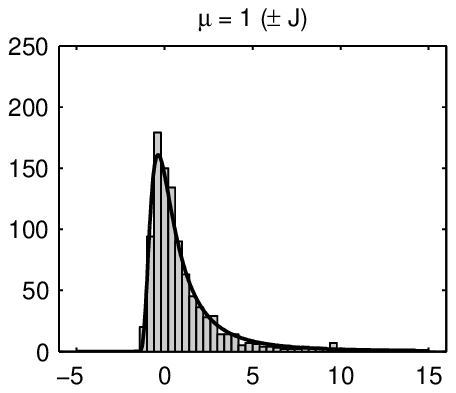,height = 1.1in}
\end{center}
\vspace*{-3ex}
\caption{Distribution of $E_G$ for the transition from $\pm J$ to Gaussian spin glasses for $n=10\times 10$.}
\label{figure-Frechet-transition}
\end{figure}

\begin{figure}[t]
\begin{center}
\subfigure[Location]{\epsfig{file=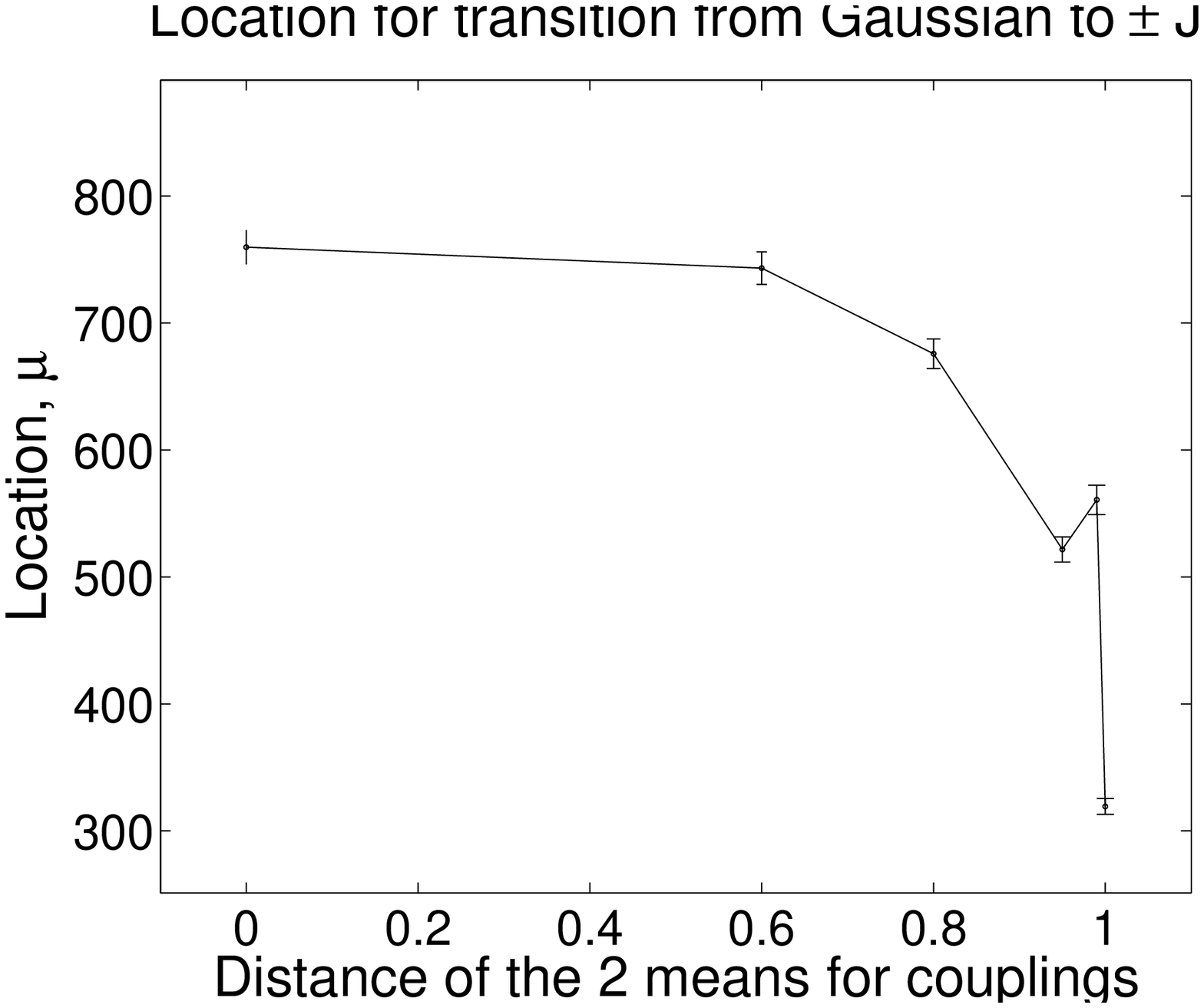,width=0.35\textwidth}}
\hspace*{0.5in}
\subfigure[Shape]{\epsfig{file=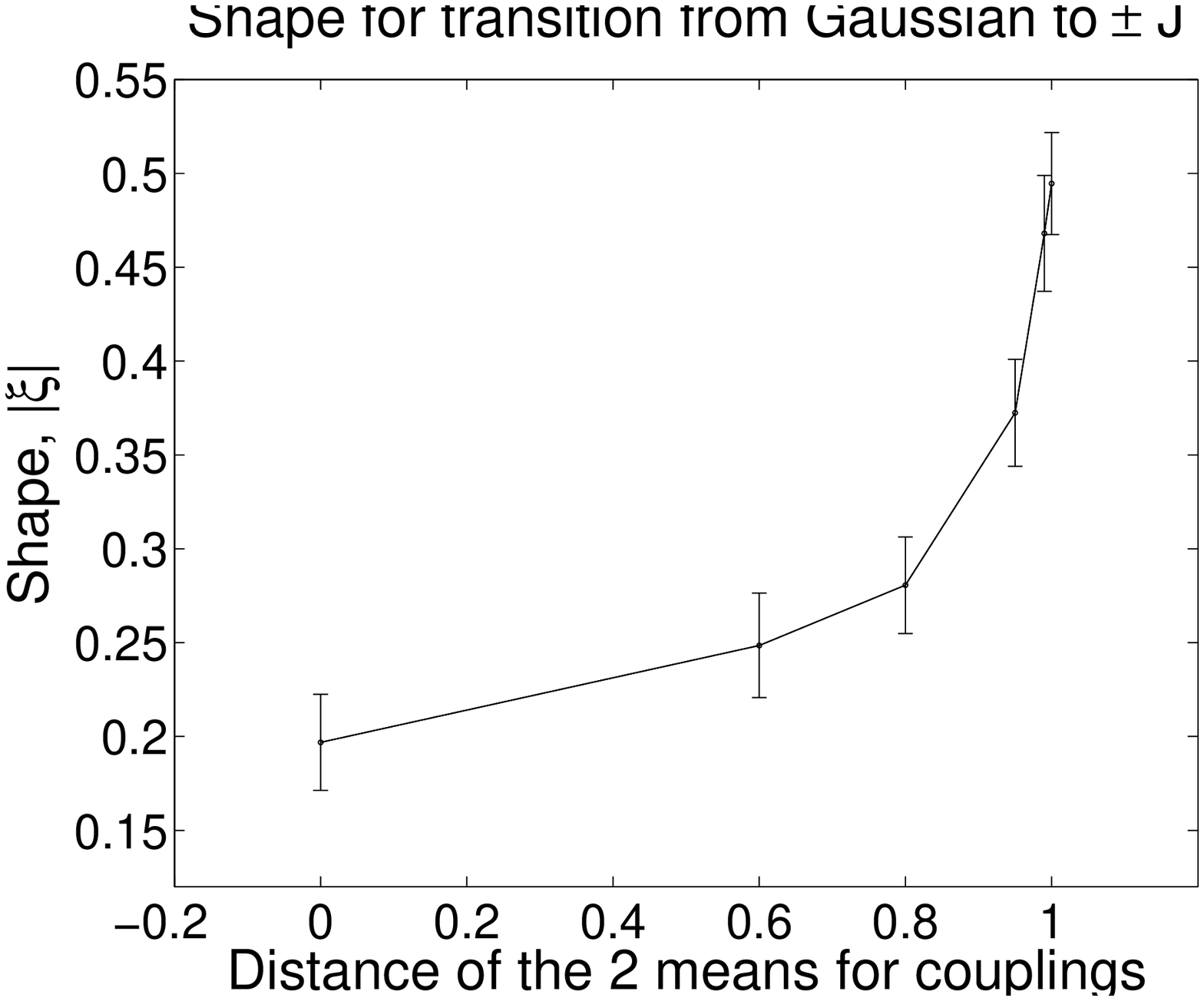,width=0.35\textwidth}}
\end{center}
\vspace*{-3ex}
\caption{Location $\mu$ and shape $\xi$ for the transition between
$\pm J$ and Gaussian spin glasses. Standard error of the estimations displayed with error bars. X-axis denotes the distance $\tilde{\mu}$ of the means used to generate couplings.}
\label{fig-location-and-shape-transition}
\end{figure}

For the transition between $\pm J$ and Gaussian couplings, $E_G$
and $E_L$ also follow Fr\'echet distributions.
Figure~\ref{figure-Frechet-transition} shows the distribution of
$E_G$ in the transition, including $\pm J$ and Gaussian cases.
Figure~\ref{fig-location-and-shape-transition} shows location and
shape parameters for the transition.

We can see that both location and shape parameters for the
transition between $\pm J$ and Gaussian couplings lie between the
corresponding parameters for the two extreme cases. That means
that considering the two extreme cases provides insight not only
in the cases themselves, but it can be used to guide estimation of
parameters for a large class of other distributions of couplings.

\section{Discussion}

In the following we discuss the experimental results, first in the
context of hBOA scalability theory and then in comparison with
flat-histogram Monte Carlo results~\cite{Dayal:04}. We close by
presenting some general conclusions for genetic and evolutionary
computation.

\subsection{Experimental results and hBOA theory}

An interesting question is whether the results obtained can be
explained using hBOA convergence theory designed for a rather
idealized situation, where the problem can be decomposed into
subproblems of bounded order over multiple levels of difficulty.
For random 2D Ising spin glasses, it can be shown that for a
complete single-level decomposition it would be necessary to
consider subproblems of order proportional to $\sqrt{n}$ as hypothesized by M\"{u}hlenbein ~\cite{Muhlenbein:99a}, which
would lead to exponentially sized populations~\cite{Pelikan:02a}.
Despite this, the number of function evaluations grows as a
low-order polynomial of the number $n$ of spins as predicted by
hBOA scalability theory for decomposable problems of bounded
difficulty~\cite{Pelikan:02a}. Spin glasses with $\pm J$ couplings
correspond to uniform scaling, where the theory predicts
$O(n^{1.55})$ evaluations; indeed the location parameter $\mu$
indeed seems to approach a polynomial of order approx. $1.5$. Spin
glasses with Gaussian couplings exhibit a non-uniform scaling,
where exponential scaling can be taken as a bounding case. For
exponential scaling, the number of evaluations would be predicted
to grow as $O(n^2)$; here the location parameter seems to grow
slightly faster with a polynomial of order approx. $2.2$. However,
the order of this polynomial decreases with problem size.

\subsection{Comparison to flat-histogram Monte Carlo}

Monte Carlo (MC) methods are usually used to integrate a function
$f(x)$ with some probability density distribution over the input
parameter $x$. The common approach is to sample a series of values
of $x$ according to the specified probability distribution, and
averaging the values of $f(x)$.

While conventional MC has been successfully used in numerous
applications, it sometimes produces inferior results for low
temperatures because the random walk through the space of all
possible configurations (values of $x$) of the system has
difficulties in overcoming energy barriers. One of the ways to
alleviate this difficulty is to modify the simulated
statistical-mechanical ensemble and use Wang-Landau
sampling~\cite{Wang:01} to sample each energy level equally
likely, thereby producing a flat histogram. The Wang-Landau
algorithm thus represents a class of methods also known as
flat-histogram MC. This approach not only alleviates the problem
of energy barriers, but it also enables computation of the number
of configurations at different energy levels, which can in turn be
used to quickly compute thermal averages for any given temperature
without having to rerun the simulation.

For flat-histogram MC, the distribution of round-trip times in
energy measured by the total number of applications of local
operators was recently shown to follow Fr\'echet distributions
\cite{Dayal:04}. However, the absolute value of the shape
parameter for flat-histogram MC was shown to approach $1$. As a
result, the mean of this distribution is not defined. Further, the
location parameter found for flat-histogram MC grows exponentially~\cite{Dayal:04}, although for this class of spin glasses it is
possible to analytically compute the entire energy spectrum in
polynomial time, $O(n^{3.5})$~\cite{Galluccio:99}.

\subsection{Important lessons for genetic and evolutionary computation}

The results presented in this paper indicate that it can be
misleading to estimate the mean convergence time by an average
over several independent samples (runs), because in some cases the
mean, variance, and other moments of the respective distribution
may become ill-defined. In this work, the location parameter
serves as a well-defined quantity to express computational
complexity of various optimization and simulation techniques,
including hBOA and flat-histogram MC. It can be expected that
similar distribution will be observed for other evolutionary
algorithms, as they reflect intrinsic properties of the spin glass
\cite{Dayal:04}.

Random 2D Ising spin glasses represent interesting classes of constraint
satisfaction problems with a large number of local optima. The results
presented in this work indicate that for such classes of problems,
recombination-based search can provide optimal solutions in low-order
polynomial time, whereas mutation-based methods scale exponentially. However,
local search is still beneficial for local improvement of solutions in
recombination-based evolutionary algorithms, because incorporating local
search decreases population sizing requirements. A similar observation was found
for MAXSAT~\cite{Pelikan:03*}.


\section{Conclusions}
\label{section-conclusions}

Random classes of Ising spin glass systems represent an
interesting class of constraint satisfaction problems for
black-box optimization. Similar to flat-histogram MC,
computational complexity of hBOA---expressed in the number of
solutions explored by both hBOA and the local hill climber until
the optimum---is found to show large sample-to-sample variations.
The obtained distribution of optimization steps follow a
fat-tailed Fr\'echet extremal value distribution. However, for
hBOA the shape parameter defining the decay of the tail is small
enough for the first two moments of the observed distributions to
exist for all but smallest system sizes. The location parameter as
well as the mean of this distribution scale like a polynomial of
low order. The experiments show that similar behavior can be
observed for $\pm J$ and Gaussian spin glasses, as well as for the
transition between these two cases. For $\pm J$ spin glasses,
performance of hBOA agrees with scalability theory for hBOA on
uniformly scaled problems, whereas for Gaussian spin glasses,
performance of hBOA agrees with scalability theory for hBOA on
exponentially scaled problems.

There are some general conclusions for genetic and evolutionary
computation. First, measuring time complexity by the average
number of function evaluations until the optimum is found can
sometimes be misleading when rare events dominate the
sample-to-sample variations. Second, it was shown for this
specific problem that recombination-based search can efficiently
deal with exponentially many local optima and still find the
global optimum in low-order polynomial time.

\section*{Acknowledgments}
\vspace*{-1ex}

Pelikan was supported by the Research Award at the University of
Missouri at St. Louis and the Research Board at the University of
Missouri. Trebst and Alet acknowledge support from the Swiss National Science
Foundation. Most of the calculations were performed on the Asgard cluster at ETH Z\"{u}rich. The hBOA software, used by Pelikan, was developed by Martin Pelikan and David E. Goldberg at the University of Illinois at Urbana-Champaign. 2D spin glass instances with ground states obtained from S. Sabhapandit and S. N. Coppersmith from the University of Wisconsin.

\vspace*{-2ex}

\begin{small}
\bibliographystyle{splncs}
\bibliography{mybib}

\begin{thebibliography}{10}

\bibitem{Mezard:86}
M{\'{e}}zard, M., Parisi, G., Virasoro, M.A.:
\newblock Spin glass theory and beyond.
\newblock World Scientific, Singapore (1987)

\bibitem{Wang:01}
Wang, F., Landau, D.P.:
\newblock Efficient, multiple-range random walk algorithm to calculate the
  density of states.
\newblock Physical Review Letters \textbf{86} (2001)  2050--2053

\bibitem{Berg:92}
Berg, B.A., Neuhaus, T.:
\newblock Multicanonical ensemble - a new approach to simulate first order
  phase-transition.
\newblock Physical Review Letters \textbf{68} (1992)

\bibitem{Pelikan:01*}
Pelikan, M., Goldberg, D.E.:
\newblock Escaping hierarchical traps with competent genetic algorithms.
\newblock Proceedings of the {G}enetic and {E}volutionary {C}omputation
  {C}onference ({GECCO}-2001) (2001)  511--518 Also {IlliGAL Report No.}
  2000020.

\bibitem{Pelikan:03b}
Pelikan, M., Goldberg, D.E.:
\newblock A hierarchy machine: {L}earning to optimize from nature and humans.
\newblock Complexity \textbf{8} (2003)

\bibitem{Chickering:97}
Chickering, D.M., Heckerman, D., Meek, C.:
\newblock A {B}ayesian approach to learning {B}ayesian networks with local
  structure.
\newblock Technical Report MSR-TR-97-07, Microsoft Research, Redmond, WA (1997)

\bibitem{Harik:95a}
Harik, G.R.:
\newblock Finding multimodal solutions using restricted tournament selection.
\newblock Proceedings of the {I}nternational {C}onference on {G}enetic
  {A}lgorithms ({ICGA}-95) (1995)  24--31

\bibitem{Pelikan:thesis}
Pelikan, M.:
\newblock Bayesian optimization algorithm: {F}rom single level to hierarchy.
\newblock PhD thesis, University of Illinois at Urbana-Champaign, Urbana, IL
  (2002)

\bibitem{Pelikan:03*}
Pelikan, M., Goldberg, D.E.:
\newblock Hierarchical boa solves ising spin glasses and maxsat.
\newblock Proceedings of the {G}enetic and {E}volutionary {C}omputation
  {C}onference ({GECCO}-2003) \textbf{{II}} (2003)  1275--1286 Also {IlliGAL
  Report No.} 2003001.

\bibitem{Fisher:28}
Fisher, R.A., Tippett, L.H.C.:
\newblock Limiting forms of the frequency distribution of the largest and
  smallest member of a sample.
\newblock In: Proceedings of Cambridge Philosophical Society. Volume~24. (1928)
   180--190

\bibitem{Galluccio:99}
Galluccio, A., Loebl, M.:
\newblock A theory of {P}faffian orientations. {I}. {P}erfect matchings and
  permanents.
\newblock Electronic Journal of Combinatorics \textbf{6} (1999) Research Paper
  6.

\bibitem{Galluccio:99a}
Galluccio, A., Loebl, M.:
\newblock A theory of {P}faffian orientations. {II}. {T}-joins, k-cuts, and
  duality of enumeration.
\newblock Electronic Journal of Combinatorics \textbf{6} (1999) Research Paper
  7.

\bibitem{Dayal:04}
Dayal, P., Trebst, S., Wessel, S., W{\"{u}}rtz, D., Troyer, M., Sabhapandit,
  S., Coppersmith, S.N.:
\newblock Performance limitations of flat histogram methods.
\newblock Physical Review Letters (in press)

\bibitem{Muhlenbein:99a}
M{\"{u}}hlenbein, H., Mahnig, T., Rodriguez, A.O.:
\newblock Schemata, distributions and graphical models in evolutionary
  optimization.
\newblock Journal of Heuristics \textbf{5} (1999)  215--247

\bibitem{Pelikan:02a}
Pelikan, M., Sastry, K., Goldberg, D.E.:
\newblock Scalability of the {B}ayesian optimization algorithm.
\newblock International Journal of Approximate Reasoning \textbf{31} (2002)
  221--258 Also {IlliGAL Report No.} 2001029.

\end{thebibliography}
\end{small}

\end{document}